\documentclass[10pt]{article} 

\usepackage[preprint]{tmlr}

\usepackage{amsmath,amsfonts,bm}

\def\eqref#1{equation~\ref{#1}}

\def\1{\bm{1}}

\DeclareMathAlphabet{\mathsfit}{\encodingdefault}{\sfdefault}{m}{sl}
\SetMathAlphabet{\mathsfit}{bold}{\encodingdefault}{\sfdefault}{bx}{n}

\usepackage{hyperref}
\usepackage{url}

\usepackage{amsfonts}       
\usepackage{nicefrac}       

\usepackage{xcolor}         

\usepackage{algorithm}

\usepackage{algorithmic}
\usepackage{amsmath}
\usepackage{booktabs}
\usepackage{graphicx}
\usepackage{amssymb}
\usepackage{multirow}

\title{STROKE: Simple Sparse Structural Occlusions for Black-Box Adversarial Attacks}

\author{
Ankit Gupta$^{1,*}$ \qquad
Akshat Pandya$^{2}$ \qquad
Paul Lin$^{1}$ \\[0.3em]
Christoph Adami$^{3,4,5}$ \qquad
Emily Dolson$^{1,3}$ \\[0.5em]
$^1$Department of Computer Science \& Engineering \qquad
$^2$Coco Robotics \\ 
$^3$Program in Evolution, Ecology, \& Behavior \qquad
$^4$Department of Physics \& Astronomy \qquad \\
$^5$Department of Microbiology, Genetics, \& Immunology \\
Michigan State University \\[0.3em]
\texttt{guptaa23@msu.edu} \\
$^*$Corresponding author
}

\def\month{MM}  
\def\year{YYYY} 
\def\openreview{\url{https://openreview.net/forum?id=XXXX}} 

\begin{document}

\maketitle


\begin{abstract}
We introduce STROKE, a query-efficient score-based black-box
adversarial attack that constructs perturbations from simple
oriented rectangular occlusions. STROKE optimizes these
perturbations through a lightweight greedy add-and-refine
procedure. Across diverse image classifiers, STROKE achieves
high attack success with low query counts at nominal 1--2\%
area settings, showing favorable performance relative to
existing patch-based attacks in several configurations.
Matched comparisons further demonstrate substantial
reductions in wall-clock attack time, while ablation studies
examine the key design choices and effectiveness across
structural primitives. These findings suggest that simple
structural occlusions offer an effective and computationally
efficient search space for black-box adversarial attacks.
\end{abstract}

\begin{figure*}[h]
\centering
\setlength{\tabcolsep}{3pt}

\begin{tabular}{cccc}
\includegraphics[width=0.30\textwidth]{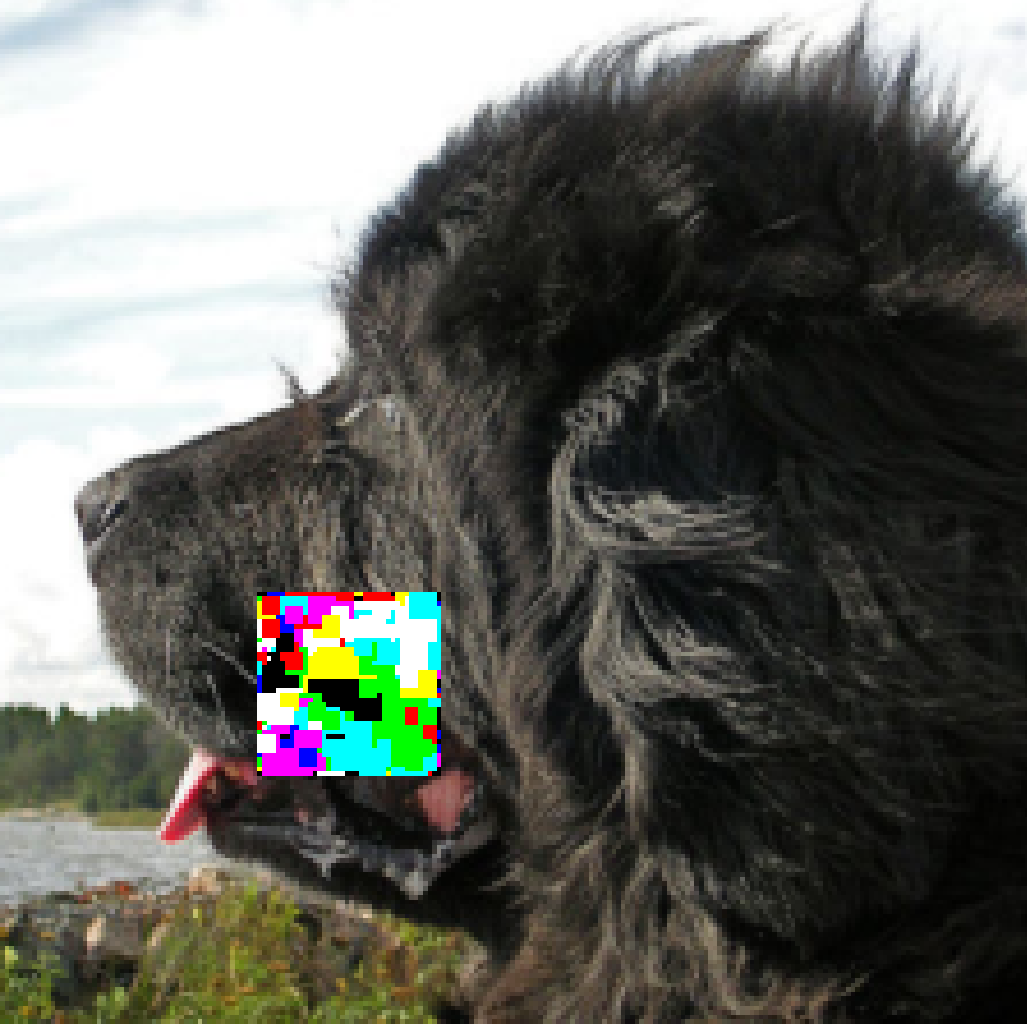} &
\includegraphics[width=0.30\textwidth]{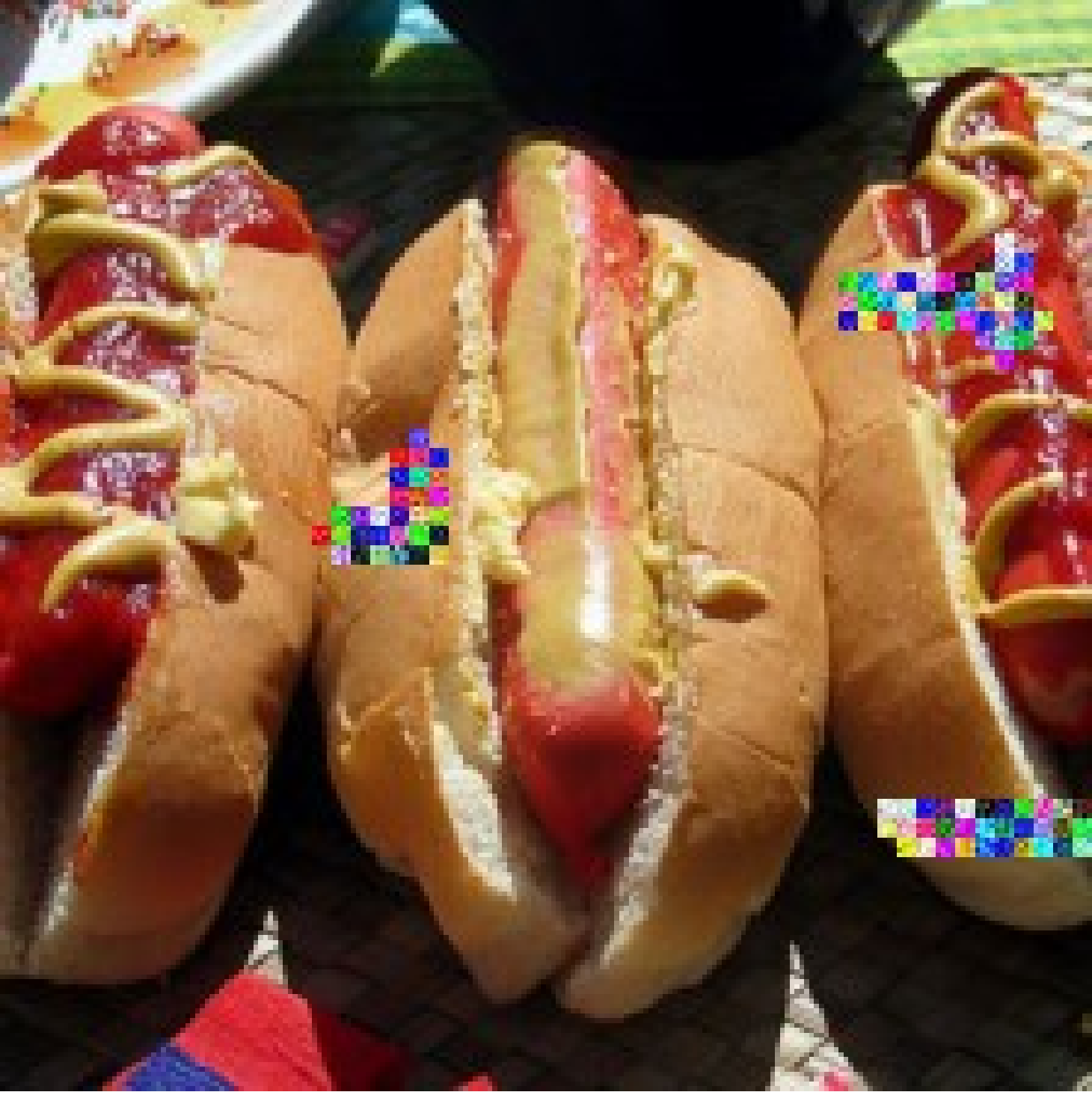} &
\includegraphics[width=0.30\textwidth]{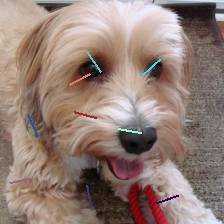} \\

\small Patch-RS \cite{croce2022sparserS} &
\small IMPACT \cite{yang2025impact} &
\small \textbf{STROKE (ours)}
\end{tabular}

\caption{
Representative adversarial perturbations produced by different black-box attacks. STROKE produces sparse structural occlusions that visually resemble simple markings or scratches. We compare their attack success, query efficiency, and runtime against the evaluated patch-based methods at the evaluated nominal area settings.
(see Tabs.~\ref{tab:attack_results} and~\ref{tab:runtime_resnet50}).
}
\label{fig:teaser}
\end{figure*}

\section{Introduction}

\label{sec:intro}
Deep neural networks have become the default backbone for visual recognition, enabling rapid progress on large-scale benchmarks such as ImageNet \cite{krizhevsky2012imagenet,russakovsky2015imagenet}. However, their remarkable accuracy can mask brittle failure modes: carefully crafted perturbations can cause confident misclassification even when inputs remain visually plausible \cite{szegedy2013intriguing,goodfellow2014explaining}. This gap between nominal performance and worst-case behavior has motivated a large literature on adversarial robustness, spanning both attack development and defenses such as adversarial training \cite{madry2018towards}.

While many early attacks assumed white-box access, realistic threat models often restrict the attacker to \emph{query-only} interaction with a deployed model (e.g., cloud APIs or locked-down inference services). In this black-box setting, the central challenge is query efficiency: each model evaluation incurs latency, monetary cost, and potential rate limits. A range of query-based attacks have therefore been proposed, including gradient-estimation, random-search strategies, and arbitrary sparse pixel perturbations \cite{chen2017zoo,guo2019simba,ilyas2018blackbox,andriushchenko2020square,Vo_Abbasnejad_Ranasinghe_2024}.

Beyond imperceptible perturbations, adversarial attacks can also operate through visible, spatially constrained modifications such as patches and occlusions \citep{brown2017adversarial,eykholt2018robust}. These settings impose explicit perturbation-area constraints, often alongside limited model-query budgets. Attack methods must therefore balance success rate, query efficiency, and computational overhead. We focus on this budgeted black-box setting, studying whether simple structural parameterizations can provide an efficient alternative to more elaborate patch optimization procedures.

Recent work has taken important steps toward query-efficient patch-style attacks under explicit area budgets. Random-search frameworks such as \emph{Sparse-RS} demonstrate that carefully designed sampling and acceptance rules can be highly effective for sparse black-box attacks \cite{croce2022sparserS}. More recently, \emph{IMPACT} proposed a flexible formulation that jointly optimizes patch mask and content using a two-phase, gradient-free procedure, achieving strong performance under strict pixel-budget constraints \cite{yang2025impact}. The method relies on evolutionary search to optimize irregular and multi-component patches. Despite these advances, a practical gap remains: increased patch expressivity and multi-factor optimization often introduce substantial computational overhead, making these methods difficult to scale across models, budgets, and evaluation settings. 

In this paper we introduce \textbf{STROKE}: Simple Sparse \textbf{Str}uctural \textbf{O}cclusions for Blac\textbf{k}-Box Adv\textbf{e}rsarial Attacks, a query-based adversarial attack designed for coverage-regulated black-box optimization. STROKE represents perturbations as a composition of simple \emph{strokes}---oriented filled rectangular occlusions with controllable length, width, location, and (optionally) color---and searches this space using a greedy \emph{add-and-refine} strategy.

The attack proceeds in two stages: it first \emph{adds} strokes to reduce confidence in the clean prediction (or increase confidence in a target class), and then \emph{refines} existing strokes by resampling their parameters while holding the remaining perturbation fixed. This decomposition supports efficient batched optimization while maintaining a compact structural representation. The resulting perturbations are visually sparse and may resemble simple markings or scratches, although their physical realizability is not assumed by the attack formulation.

At corresponding nominal area settings and query limits, we compare STROKE with IMPACT and Patch-RS and observe favorable candidate-discovery efficiency and substantial wall-clock speedups in the evaluated implementations (see Tab.~\ref{tab:attack_results}, ~\ref{tab:runtime_resnet50}, and Figs.~\ref{fig:method_sweep}, ~\ref{fig:stroke_families}).

\paragraph{Contributions.}
We summarize our main contributions as follows:
\begin{itemize}

\item \textbf{Stroke-based parameterization for sparse black-box attacks.}
We introduce STROKE, a query-based adversarial attack that represents perturbations as a composition of simple oriented rectangular occlusions.
This structured representation defines a compact search space for sparse perturbations, while remaining expressive enough to induce high-confidence misclassifications at the evaluated pixel-coverage settings.

\item \textbf{Greedy add-and-refine optimization.}
We propose a lightweight two-stage search strategy that progresses from adding new strokes to refining existing ones.
This decomposition enables efficient exploration at the evaluated coverage settings and query limits.

\item \textbf{Strong candidate-discovery efficiency.}
At the evaluated nominal area settings, STROKE achieves a strong success--query trade-off across ImageNet models, with particularly large gains on VGG-16 and ViT-B/16, and competitive performance on ResNet-50, including in the targeted setting.

\item \textbf{Substantial wall-clock efficiency in matched reruns.}
In matched reruns with identical hardware, batching, and query accounting, STROKE substantially reduces wall-clock attack time while maintaining strong attack effectiveness.

\item \textbf{Extensive empirical evaluation.}
We evaluate across diverse CNN and transformer architectures, multiple coverage settings and query limits, and additional datasets (CIFAR-10/100), and provide ablations that isolate the impact of occlusion geometry, per-occlusion proposal budget, and occlusion color.

\end{itemize}

\section{Related Work}

\paragraph{Adversarial examples and robustness.}
Modern visual recognition systems built on large-scale datasets such as ImageNet \cite{deng2009imagenet,russakovsky2015imagenet} have achieved impressive accuracy \cite{krizhevsky2012imagenet}, yet are vulnerable to adversarially crafted inputs \cite{szegedy2013intriguing,goodfellow2014explaining}. This phenomenon has motivated the development of both stronger attacks and defenses, with adversarial training remaining one of the most widely used defense paradigms \cite{madry2018towards}. Robustness evaluation also depends critically on using sufficiently strong attacks; e.g., optimization-based attacks such as C\&W revealed that several earlier defenses offered only illusory robustness \cite{carlini2017towards}.

\paragraph{Query-based black-box attacks.}
Black-box threat models restrict the attacker to querying a deployed classifier, often making query efficiency the primary bottleneck. A large body of work studies \emph{score-based} attacks that leverage confidence scores or logits, including zeroth-order optimization and randomized search methods \cite{chen2017zoo,guo2019simba,andriushchenko2020square,ilyas2018blackbox,ilyas2019bandits}. In parallel, \emph{decision-based} (hard-label) attacks assume access only to the top-1 prediction and often operate by searching along the decision boundary \cite{brendel2017decision,chen2020hopskipjump,li2020qeba,chen2020rays}. These methods highlight a general theme: carefully chosen low-dimensional search directions and parameterizations can substantially reduce query complexity relative to naive gradient estimation.

\paragraph{Location-optimized, sparse, and patch-based attacks.}
Prior work has explored location-optimized perturbations that leverage object location to construct adversarial signals \cite{giulivi2023scratches}. The \emph{adversarial patch} line of work demonstrates universal and physically realizable patch attacks \cite{brown2017adversarial,athalye2018synthesizing,eykholt2018robust,kurakin2016physical}, while other work studies visible but spatially constrained perturbations in the digital setting \cite{karmon2018lavan}. Sparse attacks that explicitly control the number of modified pixels have also been explored in both white-box and black-box contexts \cite{su2019onepixel,modas2019sparsefool,croce2022sparserS}. Most closely related to our setting are recent query-based attacks designed for pixel/area-budgeted occlusions, including patch-style random search and multi-component patch compositions \cite{croce2022sparserS,yang2025impact}.

\paragraph{Positioning.}
STROKE focuses on efficient black-box optimization in the structured sparse-occlusion setting. Building on prior work on localized patches, multi-component perturbations, and adversarial scratches \citep{croce2022sparserS,yang2025impact,giulivi2023scratches}, we adopt a compact parameterization of oriented rectangular strokes coupled with a lightweight greedy add-and-refine search. This design achieves a favorable success--query--runtime trade-off across the evaluated classifiers.

The evaluated methods differ in their perturbation geometry: Patch-RS uses a single contiguous patch, IMPACT optimizes irregular multi-component patches, and STROKE constructs perturbations from simple structural elements. Unrestricted $\ell_0$ attacks permit arbitrary sparse pixel modifications. We compare the evaluated methods at corresponding nominal area settings and query limits, with matched reruns providing a direct assessment of computational efficiency.

\section{Method: STROKE}
\label{sec:method}

We propose \textbf{STROKE}, a query-based black-box attack for image classifiers with perturbation-area control and bounded model-query access. STROKE constructs an adversarial example as a composition of a small number of simple \emph{strokes}---filled, rotated rectangles---and searches over their parameters using a greedy \emph{add-and-refine} strategy. The key design goal is to obtain a favorable \emph{success--query} trade-off at nominal coverage settings (e.g., 1\%--2\%), while keeping the inner loop lightweight enough to scale across architectures.

\subsection{Threat Model and Budgeted Occlusion Setting}
\label{sec:threat_model_method}

We consider an image classifier $f:\mathbb{R}^{H\times W\times 3}\rightarrow\mathbb{R}^C$ that outputs logits over $C$ classes, with predicted probabilities $\pi(x)=\operatorname{softmax}(f(x))$. The attacker has \emph{query access} to $f$ and aims to identify adversarial candidates through coverage-regulated stroke optimization. We study both:

\begin{itemize}
    \item \textbf{Untargeted attack.} Given clean prediction $y=\arg\max_c f_c(x)$, find $\hat{x}$ such that $\arg\max_c f_c(\hat{x})\neq y$.
    \item \textbf{Targeted attack.} Given a target $t\neq y$, find $\hat{x}$ such that $\arg\max_c f_c(\hat{x})=t$.
\end{itemize}

Let $M_{\mathcal{S}_i}\in\{0,1\}^{H\times W}$
denote the union of stroke masks applied to image $i$.
We define its normalized perturbation coverage as
\begin{equation}
\label{eq:coverage}
\rho_i = \frac{\|M_{\mathcal{S}_i}\|_0}{HW}.
\end{equation}

The transition from ADD to REFINE is determined by
perturbation coverage across the evaluated images.
The query budget counts candidate-evaluation rounds.
The complete search procedure and transition criterion
are detailed in Algorithm~\ref{alg:stroke}
in the Appendix.

\subsection{Stroke Parameterization}
\label{sec:stroke_param}
A \textbf{stroke} is defined as a filled, rotated rectangular occlusion parameterized by its center location, orientation, length and width of occlusion, and color:
\begin{equation}
\label{eq:stroke_params}
s \;=\; (x_c, y_c, \theta, \ell, w, c),
\end{equation}
where $(x_c,y_c)\in[0,W)\times[0,H)$ is the center, $\theta\in[0,2\pi)$ is the rotation angle, $\ell$ and $w$ are the stroke length and width (in pixels), and $c\in[0,1]^3$ is an RGB color. We consider two color modes: \textbf{fixed} color (e.g., white or black) and \textbf{multicolor} where $c$ is sampled per proposal.

Each stroke induces a binary mask $M_s\in\{0,1\}^{H\times W}$ that selects pixels inside the rotated rectangle. For an image $x$, applying a stroke overwrites pixels inside the mask with color $c$:
\begin{equation}
\label{eq:apply_stroke}
\big(\textsc{ApplyStroke}(x,s)\big)_{uv} =
\begin{cases}
c, & \text{if } (u,v)\in M_s,\\
x_{uv}, & \text{otherwise.}
\end{cases}
\end{equation}
For an ordered stroke list $\mathcal{S}=\langle s_1,\dots,s_K\rangle$, the composition operator applies strokes sequentially:
\begin{equation}
\label{eq:compose}
\textsc{Compose}(x,\mathcal{S}) \;=\; \textsc{ApplyStroke}(\cdots(\textsc{ApplyStroke}(x,s_1),s_2)\cdots,s_K).
\end{equation}
The overall perturbation mask is the union
\begin{equation}
\label{eq:union_mask}
M_{\mathcal{S}} \;=\; \bigvee_{k=1}^{K} M_{s_k},
\end{equation}

which determines the perturbation coverage
in Eq.~\ref{eq:coverage}.

\subsection{Objectives for Untargeted and Targeted Attacks}
\label{sec:objectives}

STROKE performs greedy search using a scalar objective evaluated via model queries. Let $\pi=\operatorname{softmax}(f(\hat{x}))$ denote the predicted probability distribution for the current candidate. 

For the \textbf{untargeted} case, we minimize the model confidence assigned to the clean class:
\begin{equation}
\label{eq:obj_untargeted}
J_{\text{un}}(\hat{x}) = \pi_y .
\end{equation}

For the \textbf{targeted} case, we maximize the confidence assigned to the target class, equivalently minimizing its negative:
\begin{equation}
\label{eq:obj_targeted}
J_{\text{tar}}(\hat{x}) = -\pi_t .
\end{equation}

These objectives follow a standard score-based strategy: each query evaluates whether a proposed stroke decreases the confidence of the clean class (untargeted) or increases the confidence of the target class (targeted), providing a useful signal without requiring gradient information.

\subsection{Greedy Add-and-Refine Search}
\label{sec:add_refine}
The search space over $\mathcal{S}$ is combinatorial: each stroke introduces continuous parameters $(x_c,y_c,\theta)$, discrete color mode choices, and interactions through the union mask and coverage control. STROKE addresses this by decomposing the search into two successive search phases:

\begin{enumerate}
    \item \textbf{Add phase (construction).}
    STROKE samples candidate strokes and retains
    the best proposal within each operation,
    progressively constructing a compact perturbation.

    \item \textbf{Refine phase (replacement).}
    Once the batch-mean coverage threshold or stroke
    limit is reached, STROKE switches permanently
    to refinement, resampling existing stroke
    parameters and accepting improvements to the
    attack objective.
\end{enumerate}

Concretely, STROKE runs for up to $Q$ candidate-evaluation rounds. The algorithm proceeds through occlusion-search operations; each operation receives a small per-occlusion-operation proposal budget $B$ (POQB). In \texttt{ADD} mode, STROKE samples $B$ candidate strokes $s$ (uniform over location and angle; fixed or random color) and evaluates the resulting candidates $\hat{x}_{\text{cand}}$ via queries. It retains the best candidate sampled within each ADD operation. In \texttt{REFINE} mode, STROKE first recomposes a \emph{base image} by removing a selected stroke and reapplying all others; it then proposes $B$ replacements for the removed stroke and greedily accepts those that improve the objective. This replacement strategy yields a coordinate-descent-like procedure over stroke parameters, while maintaining a fixed number of stored strokes.

Algorithm~\ref{alg:stroke} presents the unified procedure for both untargeted and targeted settings. The only difference is the objective function (Eq.~\ref{eq:obj_untargeted} vs.\ Eq.~\ref{eq:obj_targeted}) and the corresponding success condition. In targeted mode, we assign each image an independent random target label $t\neq y$ (uniform over the remaining classes), and declare success when the model's top-1 prediction equals $t$.

\subsection{Implementation Notes and Efficiency}
\label{sec:impl_notes}
STROKE is designed to be simple to batch and implement efficiently. We generate stroke masks using a vectorized rotation-and-threshold test and apply strokes via masked overwrites. The add-and-refine structure avoids expensive population-based optimization inside the query loop: each proposal requires only mask generation, image overwrite, normalization, and a single forward pass. In our experiments, we run on a single NVIDIA H200 GPU, enabling high-throughput batched evaluation with at most one candidate evaluation per active image in each query round.

\section{Experiments}
\label{sec:experiments}

This section describes the evaluation protocol used throughout the paper, including datasets, models, baselines, and metrics. Unless stated otherwise, all results follow the coverage-regulated attack configuration described in Section~\ref{sec:method}. See supplementary material for more details of experimental parameter settings.

\subsection{Datasets}
\label{sec:datasets}
\paragraph{ImageNet-1k.}
Our primary evaluation is conducted on ImageNet-1k validation images \cite{deng2009imagenet,russakovsky2015imagenet}. Following the common protocol for query-based evaluation in recent pixel-budget benchmarks, we construct a fixed subset of $N$  sample images ($N=500$) as in Patch-RS \cite{croce2022sparserS}. This process ensures that reported success rates correspond to genuinely induced misclassification rather than pre-existing errors. For targeted attacks, we assign each image an independent random target label uniformly sampled from the $C-1$ classes excluding the clean top-1 class.

\paragraph{CIFAR-10/100.}
To test whether the method generalizes beyond ImageNet resolution and semantics, we additionally report results on CIFAR-10 and CIFAR-100~\cite{krizhevsky2009learning}. We follow the same evaluation principles and same image sample size, i.e., $N=500$ from test set in this case.

\subsection{Models}
\label{sec:models}
We evaluate STROKE on a diverse set of standard ImageNet classifiers from both convolutional and transformer families. Specifically, we include ResNet-50 \cite{he2016resnet}, VGG-16 \cite{simonyan2015vgg}, ViT-B/16 \cite{dosovitskiy2021vit} for extensive baseline comparison. We also do model sweeps for STROKE on CNN and transformer model families, i.e., DenseNet-121 \cite{huang2017densenet} and ConvNeXt-Base \cite{liu2022convnext}, and DeiT-B \cite{touvron2021deit} and Swin-B \cite{liu2021swin}. All models use public ImageNet-1k pretrained weights and their corresponding standard preprocessing pipelines. For experiments on adversarial defender models \cite{xiang2021patchguard, croce2021robustbench, madry2018towards}, we evaluate on publicly available adversarially trained checkpoints (details provided in the supplementary material).

\subsection{Attack Configuration and Budgets}
\label{sec:attack_config}

We evaluate STROKE in both \textbf{untargeted}
and \textbf{targeted} settings using coverage
thresholds $p_{\max}\in\{1\%,2\%\}$
and candidate-query limits
$Q\in\{5{,}000,\,10{,}000\}$.

STROKE uses the stroke primitive from Sec.~\ref{sec:stroke_param} with parameters $(\ell,w)$ and a proposal budget $B$ (POQB). For the main comparisons, we fix $B=25$, set the occlusion length to $\ell=24$ for untargeted attacks and $\ell=6$ for targeted attacks, and choose the width $w$ for the nominal coverage setting (1\%: $w=1$, 2\%: $w=2$).
We vary $(\ell,w,B)$ only in controlled ablation studies (Fig.~\ref{fig:stroke_ablations}).

For targeted attacks, a random target label is sampled once per image and kept fixed throughout the attack. Each experiment is repeated with multiple random seeds and we report mean results; additional implementation details are provided in the supplementary material.

\subsection{Evaluation Metrics}
\label{sec:metrics}
We report two primary metrics, following standard practice for query-based attacks under fixed budgets:
\begin{itemize}

    \item \textbf{Attack Success Rate (ASR, \%):}
    fraction of images for which the attack finds
    a successful candidate within the query limit.
    For untargeted attacks, success means the top-1
    prediction differs from the clean label;
    for targeted attacks, success means the top-1
    prediction equals the assigned target class.

    \item \textbf{Average Queries (AQ):} average candidate-evaluation rounds until recorded success per image, with unsuccessful attacks assigned the limit $Q$. We also report median queries and success-only query statistics in supplementary material.
\end{itemize}

For the reproduced baseline experiments in
Tab.~\ref{tab:runtime_resnet50}, we additionally report
attack runtime per image in seconds.
All reported metrics are aggregated over three random seeds.

\subsection{Baselines}
\label{sec:baselines}

We compare STROKE against three established
score-based black-box attacks at corresponding
nominal area settings and query limits:

\begin{itemize}
    \item \textbf{IMPACT} \citep{yang2025impact}: a multi-component irregular patch attack using a two-phase optimization procedure.
    \item \textbf{Patch-RS} \citep{croce2022sparserS}: the patch-based instantiation of Sparse-RS, which optimizes a single contiguous patch.
    \item \textbf{TPA} \citep{yang2020patchattack}: a query-based patch attack included in the established benchmark comparison.
\end{itemize}

The evaluated methods differ in perturbation
geometry and area control. STROKE employs
coverage-triggered refinement, while the baselines
use their reported perturbation-area constraints.

\textbf{Evaluation provenance.}
Table~\ref{tab:attack_results} reports our STROKE results alongside literature-reported baseline results from IMPACT \citep{yang2025impact}, including its evaluations of Patch-RS and TPA. These are not fully matched reruns. In contrast, Table~\ref{tab:runtime_resnet50} reports our reproduced comparisons using the same evaluation subset, hardware, batching, and query-accounting protocol. Additional reproduced Patch-RS results are provided in Appendix~\ref{sec:baseline_reproduction}.

\begin{table*}[!t]
\centering
\scriptsize
\setlength{\tabcolsep}{2pt}
\renewcommand{\arraystretch}{0.92}

\caption{
Comparison of query-based attacks at nominal 1\% and 2\% area settings.
Each cell reports results at 5k/10k queries; ASR (\%) is attack success rate and AQ is average queries.
Higher ASR and lower AQ are better.
}

\label{tab:attack_results}
\begin{tabular*}{\textwidth}{@{\extracolsep{\fill}}ll*{8}{c}@{}}
\toprule
\multirow{3}{*}{\textbf{Model}}
& \multirow{3}{*}{\textbf{Method}}
& \multicolumn{4}{c}{\textbf{Untargeted Attack (5k/10k)}}
& \multicolumn{4}{c}{\textbf{Targeted Attack (5k/10k)}} \\

\cmidrule(lr){3-6} \cmidrule(lr){7-10}

& & \multicolumn{2}{c}{1\%} & \multicolumn{2}{c}{2\%}
& \multicolumn{2}{c}{1\%} & \multicolumn{2}{c}{2\%} \\

\cmidrule(lr){3-4}\cmidrule(lr){5-6}
\cmidrule(lr){7-8}\cmidrule(lr){9-10}

& & ASR & AQ & ASR & AQ & ASR & AQ & ASR & AQ \\
\midrule
\multirow{4}{*}{\textbf{ResNet50}}
& \textbf{STROKE} & \textbf{90.0}/\textbf{92.8} & \textbf{1045}/\textbf{1469} & \textbf{96.8}/\textbf{98.4} & 700/\textbf{796} & 20.2/31.8 & 4551/\textbf{8183} & 27.2/43.0 & 4335/7544 \\
& IMPACT & 87.2/90.0 & 1236/1518 & 94.2/96.4 & \textbf{676}/1044 & \textbf{24.6}/\textbf{38.4} & \textbf{4379}/8239 & \textbf{38.4}/\textbf{57.8} & \textbf{4126}/\textbf{7148} \\
& Patch-RS & 82.4/88.2 & 1360/1990 & 89.8/93.6 & 982/1408 & 7.6/10.8 & 4790/9359 & 12.2/20.0 & 4711/8885 \\
& TPA & 38.0/51.0 & 3519/5705 & 51.0/57.0 & 2807/5091 & 4.3/7.6 & 4888/9730 & 7.8/15.4 & 4781/9479 \\
\midrule
\multirow{4}{*}{\textbf{VGG16}}
& \textbf{STROKE} & \textbf{100.0}/\textbf{100.0} & \textbf{231}/\textbf{231} & \textbf{100.0}/\textbf{100.0} & \textbf{176}/\textbf{176} & \textbf{57.6}/\textbf{71.8} & \textbf{3536}/\textbf{5255} & \textbf{67.8}/\textbf{80.0} & \textbf{3138}/\textbf{4417} \\
& IMPACT & 92.8/94.6 & 862/1310 & 94.2/95.2 & 841/1069 & 16.2/27.6 & 4635/8893 & 34.6/35.4 & 4327/8295 \\
& Patch-RS & 88.6/92.4 & 1114/1562 & 92.8/94.4 & 832/1191 & 10.8/16.6 & 4704/9073 & 15.6/30.8 & 4640/8425 \\
& TPA & 42.2/56.2 & 3416/5204 & 53.6/59.8 & 2686/4777 & 5.6/9.4 & 4880/9598 & 8.2/16.8 & 4805/9321 \\
\midrule
\multirow{4}{*}{\textbf{ViT-B/16}}
& \textbf{STROKE} & \textbf{95.2}/\textbf{95.8} & \textbf{880}/\textbf{1106} & \textbf{93.4}/\textbf{95.2} & \textbf{851}/\textbf{1131} & \textbf{43.2}/\textbf{69.6} & \textbf{4059}/\textbf{6086} & \textbf{51.6}/\textbf{70.8} & \textbf{3830}/\textbf{5684} \\
& IMPACT & 85.6/88.4 & 1315/1652 & 92.8/95.0 & 880/1157 & 18.4/30.6 & 4552/8553 & 30.2/48.2 & 4311/7552 \\
& Patch-RS & 80.2/86.2 & 1457/2175 & 87.4/91.8 & 1052/1524 & 6.2/9.2 & 4825/9456 & 10.6/17.8 & 4786/9057 \\
& TPA & 35.6/48.4 & 3607/5854 & 48.2/55.2 & 2956/5237 & 3.8/6.8 & 4942/9828 & 6.2/13.6 & 4855/9553 \\
\bottomrule
\end{tabular*}

\end{table*}

\begin{table}[t]
\centering
\caption{\textbf{Attack efficiency comparison for untargeted attacks on ResNet-50} with a 10{,}000-round query limit and nominal 2\% coverage setting for STROKE.
Runtime is measured on an NVIDIA H200 GPU and reported in seconds per evaluated image.
STROKE achieves high attack success with fewer queries and substantially lower runtime than IMPACT and Patch-RS.}
\label{tab:runtime_resnet50}

\begin{tabular}{l *{3}{c} @{\hspace{1.5em}} *{3}{c} @{\hspace{1.5em}} *{3}{c}}
\toprule
& \multicolumn{3}{c}{\textbf{STROKE (ours)}} & \multicolumn{3}{c}{\textbf{IMPACT}} & \multicolumn{3}{c}{\textbf{Patch-RS}} \\
\cmidrule(lr){2-4} \cmidrule(lr){5-7} \cmidrule(lr){8-10}
& ASR $\uparrow$ & AQ $\downarrow$ & Time $\downarrow$ & ASR $\uparrow$ & AQ $\downarrow$ & Time $\downarrow$ & ASR $\uparrow$ & AQ $\downarrow$ & Time $\downarrow$ \\
\midrule
\textbf{Results} & \textbf{98.4} & \textbf{796} & \textbf{0.22} & 95.8 & 839 & 22.53 & 92.4 & 1557 & 0.77 \\
\bottomrule
\end{tabular}

\end{table}

\section{Results}
\label{sec:results}

We evaluate STROKE under the budgeted query-attack protocol in Sec.~\ref{sec:attack_config}.
Tab.~\ref{tab:attack_results} reuses baseline numbers reported by IMPACT, while Tab.~\ref{tab:runtime_resnet50} reports our matched reruns on the same NVIDIA H200 GPU with batched inference.

\subsection{ImageNet-1k results}

\paragraph{Strong untargeted candidate discovery across area settings.}
At the evaluated 1--2\% area settings,
STROKE achieves strong success--query performance
relative to the reported patch-based baselines
across the three architectures in
Tab.~\ref{tab:attack_results}.
The gains are largest on VGG-16 and ViT-B/16.

\paragraph{Significantly faster attacks without sacrificing success–query efficiency.}
STROKE's greedy add-and-refine loop keeps per-query overhead low.
In the ResNet-50 runtime comparison at the nominal 2\% setting and 10k query limit (Tab.~\ref{tab:runtime_resnet50}), STROKE reaches 98.4\% ASR with 796 AQ in 0.22\,s/image, compared with 22.53\,s/image for IMPACT and 0.77\,s/image for Patch-RS.
This speedup reflects the lightweight stroke search relative to population-based patch optimization and makes large sweeps more practical.

\begin{figure*}[t]
    \centering
    \begin{minipage}{0.48\textwidth}
        \centering
        \includegraphics[width=\linewidth]{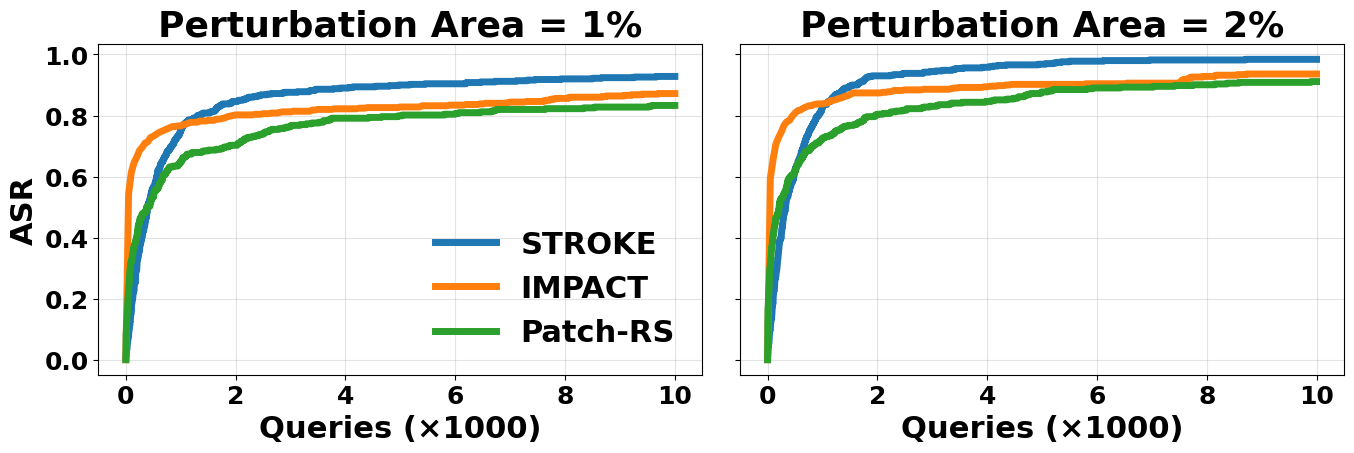}
        \caption{\textbf{ASR comparison at nominal area settings on ResNet-50.} 
        Attack success rate (ASR) versus query budget on ImageNet for STROKE, IMPACT, and Patch-RS at nominal $1\%$ and $2\%$ area settings.}
        \label{fig:method_sweep}
    \end{minipage}
    \hfill
    \begin{minipage}{0.48\textwidth}
        \centering
        \includegraphics[width=\linewidth]{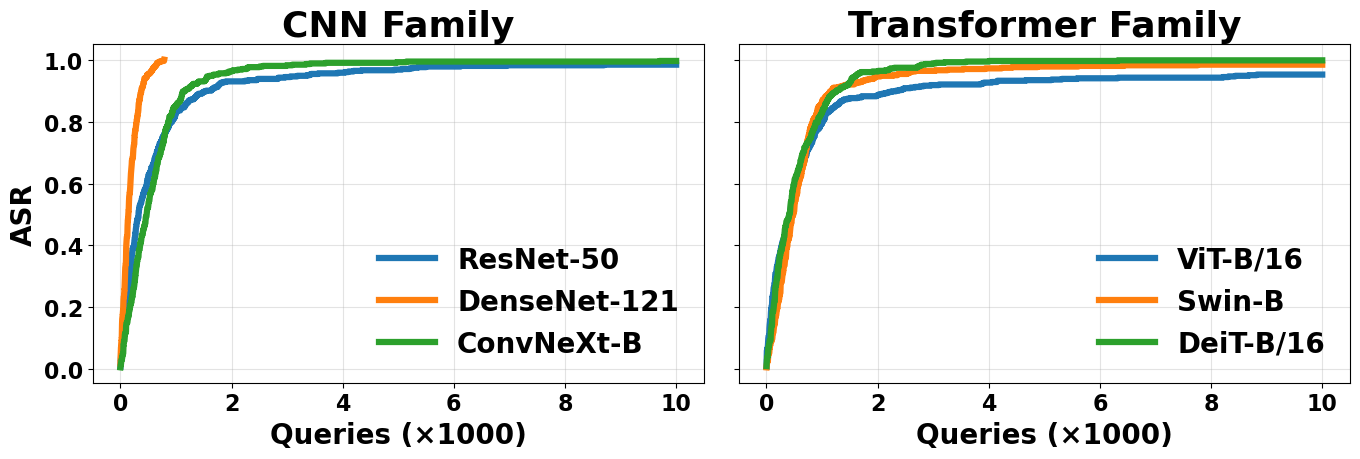}
        \caption{\textbf{Attack success rate (ASR) vs. Queries for STROKE.} 
        Results at a nominal 2\% pixel-coverage threshold for CNNs (ResNet-50, DenseNet-121, ConvNeXt-B) and Transformers (ViT-B/16, Swin-B, DeiT-B/16). STROKE achieves high success rates within a few thousand queries.}
        \label{fig:stroke_families}
    \end{minipage}
\end{figure*}

\begin{figure*}[!t]
    \centering
    \includegraphics[width=.99\linewidth]{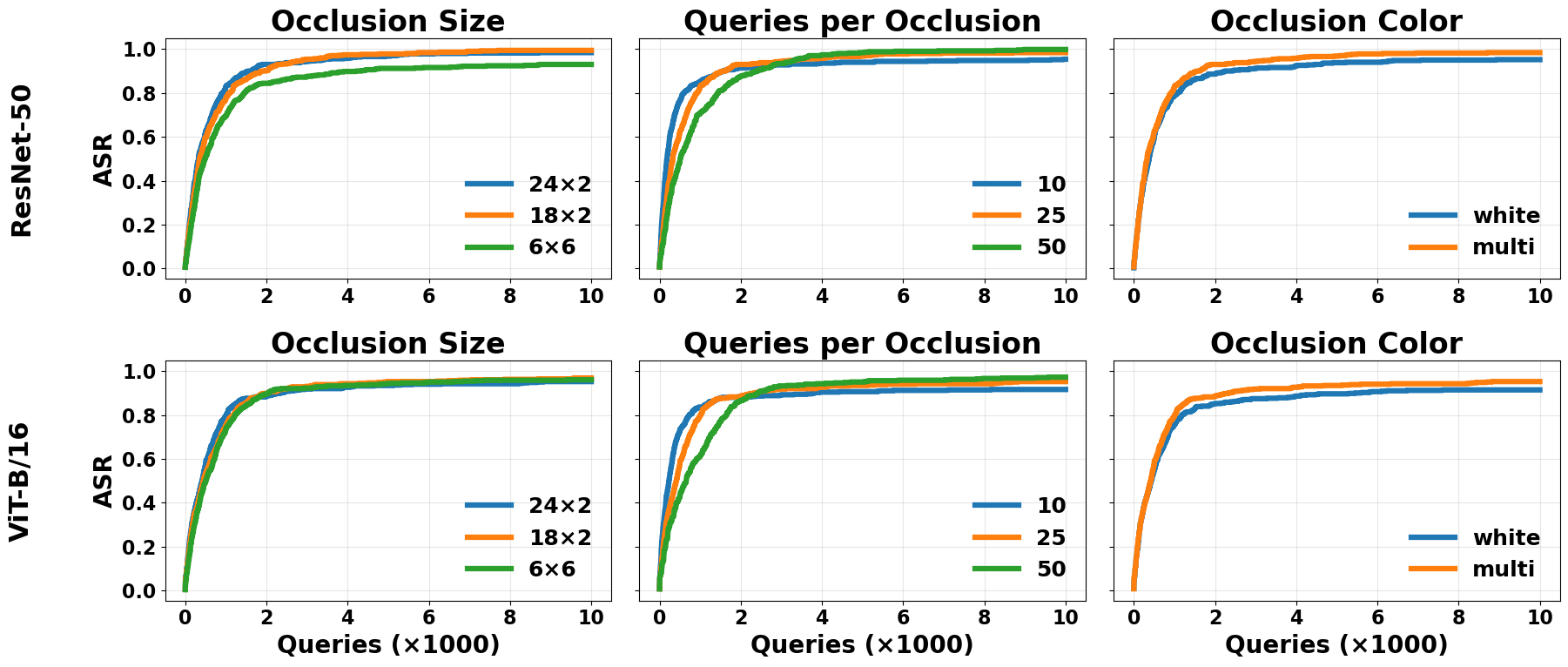}
    \caption{\textbf{STROKE ablations at a nominal 2\% coverage threshold and 10k-query limit (untargeted).}
    ASR versus queries on ImageNet for \textbf{ResNet-50} (top row) and \textbf{ViT-B/16} (bottom row).
    We ablate (left) occlusion size ($\ell$ x $w$) for POQB = 25, color = multi, (middle) queries per occlusion (POQB) for size = 24x2, color = multi, and (right) occlusion color mode for POQB = 25, size = 24x2.}
    \label{fig:stroke_ablations}
\end{figure*}

\paragraph{Early attack success--query trade-off.}
Beyond endpoint ASR, Fig.~\ref{fig:method_sweep} shows that STROKE reaches high success earlier in the query budget at both nominal 1\% and 2\% coverage settings.

\paragraph{Targeted attacks: strong gains on ViT-B/16 and VGG-16.}
In targeted attacks, STROKE gives the largest gains on ViT-B/16 and VGG-16, improving both success and query efficiency at nominal 1--2\% coverage settings (Tab.~\ref{tab:attack_results}).
On ResNet-50, it remains competitive but trails the strongest baseline in some targeted configurations, consistent with the harder targeted search regime.

\paragraph{Broad generalization across CNN and transformer backbone family.}
Fig.~\ref{fig:stroke_families} shows that STROKE achieves high ASR within a few thousand queries across additional CNNs and transformers with a nominal 2\% coverage threshold, suggesting that sparse structural occlusions expose a shared vulnerability.

\paragraph{What matters for performance: Ablation.}
The ablations in Fig.~\ref{fig:stroke_ablations} show that moderately sized strokes balance coverage and controllability, larger proposal budgets improve ASR with diminishing query returns, and multicolor strokes consistently outperform fixed-color strokes. 
We further evaluate approximately area-matched elliptical and circular primitives (Appendix~\ref{sec:primitive_ablation}). These alternatives also achieve strong attack performance, indicating that the effectiveness of the structured search is not specific to rectangular strokes.


\paragraph{Contribution of refinement.}
To isolate the effect of the refinement stage, we compare the complete ADD+REFINE procedure against an ADD-only variant on ResNet-50. Both use $18\times2$ strokes under a 10k-query budget. The ADD-only variant terminates at the construction-stage stopping criterion, whereas the full method continues with stroke-replacement operations.

\begin{table}[t]
\centering
\caption{\textbf{Effect of refinement on ImageNet (ResNet-50).}
Untargeted attack success rate (ASR, \%) with
$18\times2$ strokes and a 10k-query budget.}
\label{tab:refine_ablation}
\small
\begin{tabular}{lccc}
\toprule
\textbf{Coverage threshold} &
\textbf{ADD only} &
\textbf{ADD+REFINE} &
\textbf{Gain (pp)} \\
\midrule
1\% & 64.0 & \textbf{84.6} & +20.6 \\
2\% & 95.6 & \textbf{99.4} & +3.8 \\
\bottomrule
\end{tabular}
\end{table}

As shown in Table~\ref{tab:refine_ablation}, refinement improves ASR by 20.6 percentage points at the 1\% coverage threshold and 3.8 points at 2\%. The larger gain at the lower threshold illustrates the benefit of replacement operations after the construction stage.

\paragraph{Additional analyses and evaluations (see supplementary material).}
The supplementary material includes defended-model experiments, Grad-CAM and t-SNE diagnostics, transferability studies, and physical-world experiments.

\begin{table}[t]
\centering
\footnotesize
\setlength{\tabcolsep}{8pt}
\renewcommand{\arraystretch}{0.95}

\caption{
\textbf{Performance on CIFAR-10 and CIFAR-100 using ResNet-50.}
Results are reported under a query budget of 5,000 and a nominal coverage threshold of 5\% for STROKE.
ASR (\%) denotes attack success rate and AQ is average number of queries.
Higher ASR and lower AQ indicate better attack performance. Best values are in bold.
}
\label{tab:cifar_results}
\centering
\small
\begin{tabular}{l cc cc}
\toprule
\multirow{2}{*}{\textbf{Method}} & \multicolumn{2}{c}{\textbf{Untargeted (CIFAR-10/100)}} & \multicolumn{2}{c}{\textbf{Targeted (CIFAR-10/100)}} \\
\cmidrule(lr){2-3} \cmidrule(lr){4-5}
& AQ $\downarrow$ & ASR (\%) $\uparrow$ & AQ $\downarrow$ & ASR (\%) $\uparrow$ \\
\midrule
\textbf{STROKE} (ours) & \textbf{56.16} / \textbf{45.99} & \textbf{100.0} / \textbf{100.0} & \textbf{185.60} / \textbf{1551.00} & \textbf{100.0} / 80.8 \\
IMPACT   & 113.98 / 72.81 & 99.2 / 99.3 & 1634.36 / 1556.17 & 92.7 / \textbf{90.1} \\
Patch-RS & 156.78 / 98.52 & 97.8 / 98.1 & 2155.47 / 2310.80 & 85.3 / 82.5 \\
\bottomrule
\end{tabular}

\end{table}

\subsection{CIFAR-10/100 results}

Tab.~\ref{tab:cifar_results} evaluates CIFAR-10/100 at a nominal 5\% area setting and 5k-query limit.
STROKE is near-ceiling on CIFAR-10 and remains strong for untargeted CIFAR-100; targeted CIFAR-100 is harder and can trail the strongest baseline.

\subsection{Limitations}
STROKE is a \emph{score-based} attack and assumes access to confidence scores or logits (Sec.~\ref{sec:threat_model_method}).
Targeted performance is not uniformly best across all settings in Tabs.~\ref{tab:attack_results} and~\ref{tab:cifar_results}, and richer occlusion primitives may improve targeted reliability under tight budgets.

\section{Conclusion}
We introduced STROKE, a simple structural parameterization and greedy add-and-refine search procedure for coverage-regulated score-based black-box attacks. Across CNN and transformer classifiers on ImageNet-1k, STROKE achieves strong attack success and query efficiency at the evaluated area settings, with additional evaluations on CIFAR-10/100. Matched implementations demonstrate substantial runtime advantages over the evaluated patch-based baselines, while an add-only ablation establishes the contribution of refinement under tight perturbation budgets. These results support simple structured occlusions as an efficient search space for black-box adversarial optimization. Extensions to other attack settings, including hard-label feedback and more demanding physical conditions, remain directions for future work.

\section*{Statement of Broader Impact}

This work investigates vulnerabilities of image classifiers to sparse structural perturbations. By enabling more efficient robustness evaluation, STROKE may help identify weaknesses in visual recognition systems and inform the development of stronger defenses. At the same time, the proposed attack could be misused to induce incorrect predictions in deployed models. We therefore emphasize its role as a robustness evaluation method and the importance of testing appropriate defenses before deployment in safety-critical settings.

\clearpage

\bibliography{main}
\bibliographystyle{tmlr}

\appendix

\newpage

\section{Technical appendices and supplementary material}

\subsection{Additional Experimental Details}

This section provides additional implementation details and experimental configuration necessary for reproducibility. All experiments follow the evaluation protocol described in the main paper unless stated otherwise.

\subsubsection{Hardware and Runtime Environment}

All experiments were conducted on a single NVIDIA H200 GPU using PyTorch. 
Inference was performed using batched evaluation to maximize throughput while maintaining correct query accounting, where each forward pass of the model for a given image is counted as one query. Mixed-precision inference (bfloat16) was used during model evaluation to improve throughput on GPU hardware. 

\subsubsection{Dataset Construction}

ImageNet experiments follow the protocol described in the main paper. We use a fixed subset of $N=500$ validation images that are correctly classified by the clean model. Images are preprocessed using the standard transformation pipeline associated with each pretrained model (e.g., resizing, cropping, and normalization). CIFAR-10 and CIFAR-100 experiments follow the same evaluation principle using $N=500$ randomly sampled test images.

\subsubsection{Model Details}

We evaluate a range of standard ImageNet classifiers from both convolutional and transformer families using publicly available pretrained checkpoints. Models are loaded from the torchvision model zoo where available, including ResNet-50, VGG-16, ViT-B/16, DenseNet-121, ConvNeXt-B, and Swin-B. All models use their corresponding default preprocessing pipelines provided by torchvision. Transformer variants based on DeiT are loaded using the timm library with ImageNet-1k pretrained weights.

\subsubsection{Attack Hyperparameters}

Unless otherwise specified, STROKE uses the settings described in the main experiments: $Q\in\{5{,}000,10{,}000\}$ candidate-evaluation rounds and nominal coverage thresholds $p_{\max}\in\{1\%,2\%\}$. The occlusion primitive is a rotated rectangular region with length $\ell$ and width $w$, where the dimensions are chosen to be compatible with the corresponding coverage setting. During each occlusion search operation, a fixed number of candidate proposals are evaluated using a per-occlusion proposal budget (POQB) $B$. 
Each candidate is defined by sampling the rectangle parameters and evaluating the resulting perturbed image via a model query. The occlusion primitive is parameterized by its center location, orientation, and color, as described in the main paper. We consider two color modes: a fixed-color setting (e.g., white) and a multicolor setting in which RGB values are sampled uniformly for each proposal. All experiments are repeated using fixed random seeds to ensure reproducibility of the reported results.

\subsubsection{Implementation Details}

The attack is implemented in PyTorch and operates directly in pixel space. Stroke masks are generated using a batched geometric test that constructs rotated rectangular masks given a center location, orientation, length, and width. Candidate masks are applied to images using vectorized tensor operations.

For efficiency, images are stored in pixel space and converted to normalized tensors only when performing model evaluation. During each query step, candidate perturbations are generated for all currently active images in parallel, allowing batched inference across the batch of images being attacked.

The implementation maintains per-image state including the current perturbation mask, accumulated strokes, and query statistics. These statistics are logged during the attack to record quantities such as query counts, attack success rate, sparsity level, and number of strokes used.

\begin{algorithm}[!t]
\caption{\textsc{Stroke}}
\label{alg:stroke}
\begin{algorithmic}[1]
    \renewcommand{\algorithmicrequire}{\textbf{In:}}
    \renewcommand{\algorithmicensure}{\textbf{Out:}}
    \REQUIRE Class-probability model $f$,
        batch $x=\{x_i\}_{i=1}^{N}$,
        query-round budget $Q$,
        nominal coverage threshold $p_{\max}$,
        stroke size $(\ell,w)$, POQB $B$,
        max strokes $K_{\max}$,
        attack type $\mathcal A$, color policy $\mathcal C$,
        targeted early-stop flag $E$;
        $N,B,K_{\max}\ge1$, $Q\ge0$
    \ENSURE Final candidates $\hat{x}$

    \STATE $\hat{x}\gets x,\ P\gets f(x),\
        \mathcal S_i\gets\langle\rangle,\ M_i\gets0,\
        y_i\gets\arg\max_c P_{i,c}\quad(\forall i)$
    \STATE Sample $t_i\neq y_i$ uniformly if targeted;
        otherwise $t_i\gets\bot$
    \STATE $\operatorname{Obj}_i(P)\gets
        (-P_{t_i}\text{ if targeted; }P_{y_i}\text{ otherwise})$
    \STATE $\operatorname{Success}_i(P)\gets
        (\arg\max_c P_c=t_i\text{ if targeted; }
        \arg\max_c P_c\neq y_i\text{ otherwise})$
    \STATE $q\gets0,\ op\gets0,\ k\gets0,\
        r\gets\text{false},\ \mathcal V\gets\{1,\ldots,N\}$

    \WHILE{$q<Q$ \textbf{and} $\mathcal V\neq\emptyset$}
        \STATE $\mathcal I\gets\mathcal V$
            \COMMENT{Fixed batch within operation}
        \STATE $a\gets(\neg r\textbf{ and }k<K_{\max})$
        \IF{$a$}
            \STATE $j\gets k;\
                (\hat{x}_{b,i},M_{b,i})
                \gets(\hat{x}_i,M_i)
                \quad(i\in\mathcal I)$ \COMMENT{ADD}
            \STATE $\text{best}_i\gets+\infty
                \quad(i\in\mathcal I)$
                \COMMENT{Best sampled addition}
        \ELSE
            \STATE $r\gets\text{true};\
                j\gets op\bmod k$ \COMMENT{REFINE}
            \STATE $(\hat{x}_{b,i},M_{b,i})
                \gets\operatorname{Replay}(x_i,\mathcal S_i
                \text{ excluding slot }j)
                \quad(i\in\mathcal I)$
                \COMMENT{Replay in stored order}
            \STATE $\text{best}_i\gets\operatorname{Obj}_i(P_i),\
                s_i^\star\gets\mathcal S_i[j]
                \quad(i\in\mathcal I)$
        \ENDIF

        \STATE $L\gets\min(B,Q-q)$
        \FOR{$\tau=1,\ldots,L$}
            \STATE $q\gets q+1$;
                \textbf{if} $\mathcal V=\emptyset$
                \textbf{then break}
            \STATE Sample rotated strokes $s_i$ with size
                $(\ell,w)$ and color policy $\mathcal C$
                \quad$(i\in\mathcal I)$
            \STATE $\hat{x}_{c,i}\gets
                \operatorname{ApplyStroke}(\hat{x}_{b,i},s_i),\
                M_{c,i}\gets M_{b,i}\lor M_{s_i}
                \quad(i\in\mathcal I)$
            \STATE $P_c\gets f(\hat{x}_c)$
            \STATE $\mathcal J\gets
                \{i\in\mathcal I:
                \operatorname{Obj}_i(P_{c,i})<\text{best}_i\}$
            \STATE $\text{best}_i\gets
                \operatorname{Obj}_i(P_{c,i});\
                (\hat{x}_i,M_i,P_i,s_i^\star)
                \gets(\hat{x}_{c,i},M_{c,i},P_{c,i},s_i)
                \quad(i\in\mathcal J)$
            \STATE \textbf{if} untargeted \textbf{or} $E$
                \textbf{then}
                $\mathcal V\gets\mathcal V\setminus
                \{i\in\mathcal I:\operatorname{Success}_i(P_i)\}$
        \ENDFOR

        \STATE $\mathcal S_i[j]\gets s_i^\star
            \quad(i\in\mathcal I)$
            \COMMENT{Append or replace slot}
        \IF{$a$}
            \STATE $k\gets k+1;\
                \bar{\rho}\gets
                \frac{1}{N}\sum_{i=1}^{N}\frac{\|M_i\|_0}{HW}$
            \STATE $r\gets r\lor(\bar{\rho}\ge p_{\max})
                \lor(k\ge K_{\max})$
        \ENDIF
        \STATE $op\gets op+1$
    \ENDWHILE
    \RETURN $\hat{x}$
\end{algorithmic}
\end{algorithm}

\subsection{Extended Experimental Results}

\subsubsection{Additional Model Families}

To evaluate the generality of STROKE across different architecture families, we additionally test the attack on several widely used ImageNet models beyond those included in the main comparison. These include DenseNet-121 and ConvNeXt-B from the CNN family, and Swin-B and DeiT-B from the transformer family.

\begin{table}[!t]
\centering
\small
\setlength{\tabcolsep}{6pt}
\renewcommand{\arraystretch}{1.1}

\caption{\textbf{Additional architecture evaluation for STROKE.}
Results on ImageNet at a nominal 2\% coverage threshold and 10k-query limit (untargeted). We report attack success rate (ASR) and average queries (AQ).}
\label{tab:stroke_additional_models}

\begin{tabular}{lcc}
\toprule
Model & ASR (\%) $\uparrow$ & AQ $\downarrow$ \\
\midrule
DenseNet-121 & 100.0 & 180 \\
ConvNeXt-B   & 99.6  & 653 \\
Swin-B       & 98.4  & 780 \\
DeiT-B       & 99.8  & 592 \\
\bottomrule
\end{tabular}

\end{table}

Table~\ref{tab:stroke_additional_models} shows that STROKE remains highly effective across a diverse set of architectures. In particular, the attack achieves near-perfect success rates on all evaluated models while requiring relatively few queries on average. The behavior is consistent across both convolutional and transformer architectures, suggesting that sparse structural occlusions expose a broadly shared vulnerability among modern vision models.


\subsection{Additional Ablation Studies}

This section provides additional ablation studies to better understand the design choices behind STROKE. Unless otherwise stated, all experiments follow the same evaluation protocol used in the main paper: ImageNet validation images, a nominal 2\% coverage threshold, and a 10k-query limit in the untargeted setting. We report attack success rate (ASR) and average queries (AQ), where unsuccessful runs are counted as using the full query budget. (See Table~\ref{tab:stroke_ablations_supp})

\subsubsection{Occlusion Size}

We first study the effect of the occlusion geometry by varying the rectangle dimensions while keeping the proposal budget per occlusion (POQB) fixed at 25 and using multicolor strokes. As shown in Table~\ref{tab:stroke_ablations_supp}, moderately elongated strokes provide the best trade-off between coverage and controllability.

Among the evaluated rectangular configurations, the $18\times2$ setting achieves the highest attack success rate on both models while maintaining competitive query efficiency.
In contrast, the $6\times6$ configuration produces noticeably worse query efficiency on ResNet-50, suggesting that shorter, thicker occlusions reduce the effectiveness of the greedy search procedure. Overall, these results indicate that elongated occlusions allow the attack to cover semantically meaningful regions more efficiently at the evaluated coverage setting.

\subsubsection{Queries per Occlusion}

Next, we vary the per-occlusion proposal budget (POQB), which determines how many candidate strokes are evaluated before committing to the best update. Increasing the proposal budget improves the success rate slightly but introduces diminishing returns in query efficiency. 

For example, increasing POQB from 25 to 50 yields a marginal improvement in ASR but increases the average number of queries required. Conversely, reducing POQB to 10 decreases both ASR and efficiency. These results suggest that a moderate proposal budget strikes the best balance between exploration of the search space and query efficiency.

\subsubsection{Occlusion Color}

Finally, we compare fixed-color occlusions against multicolor occlusions. Allowing the attack to sample colors uniformly across proposals consistently improves performance across both models. Multicolor strokes lead to higher ASR and lower average query counts compared to fixed white occlusions.

This behavior suggests that color diversity expands the effective search space of the attack, enabling the algorithm to discover more effective perturbations at the same nominal coverage threshold and query limit.

\subsubsection{Alternative Structural Primitives}
\label{sec:primitive_ablation}

To examine whether the effectiveness of STROKE depends specifically on rectangular geometry, we additionally evaluate approximately area-matched elliptical and circular occlusions. All configurations use a nominal 2\% coverage threshold, 10k candidate-query rounds, a per-occlusion proposal budget of 25, and multicolor proposals.

\begin{table}[t]
\centering
\caption{\textbf{Comparison of structural primitives.}
Untargeted ImageNet results ($N=500$) at a nominal 2\% coverage setting and 10k-query limit. ASR is attack success rate (\%) and AQ is average queries.}
\label{tab:primitive_comparison}
\small
\begin{tabular}{lrrrr}
\toprule
& \multicolumn{2}{c}{ResNet-50}
& \multicolumn{2}{c}{ViT-B/16} \\
\cmidrule(lr){2-3}
\cmidrule(lr){4-5}
\textbf{Primitive} & ASR & AQ & ASR & AQ \\
\midrule
Rectangle ($18\times2$) & \textbf{99.4} & 798.5 & 96.8 & 1066.5 \\
Circle ($6.8\times6.8$) & 96.8 & \textbf{741.9} & 96.6 & 744.3 \\
Ellipse ($9.6\times4.8$) & 95.6 & 889.8 & 97.4 & \textbf{717.0} \\
Ellipse ($13.6\times3.4$) & 95.2 & 795.0 & \textbf{97.6} & 756.1 \\
\bottomrule
\end{tabular}
\end{table}

Table~\ref{tab:primitive_comparison} shows that strong attack performance is not restricted to rectangular strokes. While the rectangular configuration achieves the highest ASR on ResNet-50, elliptical alternatives obtain comparable success rates and favorable query efficiency, particularly on ViT-B/16. These findings suggest that simple low-dimensional structural primitives provide effective search spaces, with the preferred geometry depending on the model and operating point.

\begin{table}[!t]
\centering
\small
\setlength{\tabcolsep}{6pt}
\renewcommand{\arraystretch}{1.15}

\caption{\textbf{Ablation summary for STROKE (untargeted).}
Results on ImageNet at a nominal 2\% coverage threshold and 10k-query limit. We report attack success rate (ASR) and average queries (AQ), where unsuccessful runs are counted as using the full query budget. Best values within each ablation block are bolded.}
\label{tab:stroke_ablations_supp}

\begin{tabular}{l cc cc}
\toprule
\multirow{2}{*}{\textbf{Setting}} &
\multicolumn{2}{c}{\textbf{ResNet-50}} &
\multicolumn{2}{c}{\textbf{ViT-B/16}} \\
\cmidrule(lr){2-3}\cmidrule(lr){4-5}
& \textbf{ASR (\%) $\uparrow$} & \textbf{AQ $\downarrow$}
& \textbf{ASR (\%) $\uparrow$} & \textbf{AQ $\downarrow$} \\
\midrule

\multicolumn{5}{l}{\textbf{Occlusion Size} (POQB=25, color=multi)} \\
\quad 24$\times$2 & 98.4 & \textbf{796.1} & 95.2 & 1130.9 \\
\quad 18$\times$2 & \textbf{99.4} & 798.5 & \textbf{96.8} & \textbf{1066.5} \\
\quad 6$\times$6  & 93.0 & 1484.9 & 96.0 & 1151.0 \\
\midrule

\multicolumn{5}{l}{\textbf{Queries per Occlusion} (size=24$\times$2, color=multi)} \\
\quad 10 & 95.4 & 913.2 & 91.6 & 1238.7 \\
\quad 25 & 98.4 & \textbf{796.1} & 95.2 & \textbf{1130.9} \\
\quad 50 & \textbf{99.8} & 947.8 & \textbf{97.2} & 1235.1 \\
\midrule

\multicolumn{5}{l}{\textbf{Occlusion Color} (size=24$\times$2, POQB=25)} \\
\quad white & 95.2 & 1124.9 & 91.4 & 1489.8 \\
\quad multi & \textbf{98.4} & \textbf{796.1} & \textbf{95.2} & \textbf{1130.9} \\
\bottomrule
\end{tabular}

\end{table}

\subsection{Baseline Reproduction and Extended Runs}
\label{sec:baseline_reproduction}

The baseline numbers reported in Table 1 of the main paper are taken from the IMPACT paper for their own experiments as well as for Patch-RS and TPA. To verify these results, we reproduced several baseline runs using the publicly available implementations.

During this process, we additionally evaluated Patch-RS across multiple architectures using comparable model and query-limit configurations. Table~\ref{tab:patchrs_reproduced} reports these reproduced results for a 10k query budget under both 1\% and 2\% perturbation area constraints. The results are broadly consistent with the numbers reported in prior work while extending the evaluation to additional architectures.

\begin{table}[!t]
\centering
\small
\setlength{\tabcolsep}{5pt}
\renewcommand{\arraystretch}{1.15}

\caption{\textbf{Reproduced and extended Patch-RS baseline results.}
We reproduce Patch-RS using the publicly available implementation under its original pixel-budgeted evaluation setting. Results are reported for a 10k query budget under both 1\% and 2\% perturbation area constraints.}

\label{tab:patchrs_reproduced}

\begin{tabular}{l l cc cc}
\toprule
\multirow{2}{*}{Model} & \multirow{2}{*}{Method} &
\multicolumn{2}{c}{1\% Budget (10k)} &
\multicolumn{2}{c}{2\% Budget (10k)} \\
\cmidrule(lr){3-4}\cmidrule(lr){5-6}
 & & ASR (\%) $\uparrow$ & AQ $\downarrow$ & ASR (\%) $\uparrow$ & AQ $\downarrow$ \\
\midrule

ResNet50
& Patch-RS & 83.29 & 2406.2 & 92.43 & 1557.3 \\

ViT-B/16
& Patch-RS & 82.09 & 3588.9 & 94.28 & 2915.2 \\

DeiT-B
& Patch-RS & 83.05 & 4107.1 & 95.64 & 3606.5 \\

DenseNet-121
& Patch-RS & 88.89 & 1990.9 & 96.48 & 1386.6 \\

\bottomrule
\end{tabular}

\end{table}

\subsection{Performance on Adversarial Defenders}

To further evaluate the robustness of STROKE under defensive training and patch-based protection mechanisms, we additionally test the attack against several commonly used defender models. Specifically, we evaluate STROKE on AT-ResNet50-L2 ($\epsilon=3.0$), AT-ResNet50-Linf ($\epsilon=4/255$), and BagNet17 combined with PatchGuard. These models represent different robustness strategies, including adversarial training and architectural defenses against patch-style attacks. STROKE is evaluated untargetedly on ImageNet at a nominal 2\% coverage threshold and 10k-query limit.

\begin{table}[!t]
\centering
\small
\setlength{\tabcolsep}{6pt}
\renewcommand{\arraystretch}{1.1}

\caption{\textbf{Performance of STROKE against adversarial defender models.}
Results are reported at a nominal 2\% coverage threshold and 10k-query limit in the untargeted setting.}
\label{tab:defender_results}

\begin{tabular}{lcc}
\toprule
Model & ASR (\%) $\uparrow$ & AQ $\downarrow$ \\
\midrule
AT-ResNet50-L2 ($\epsilon = 3.0$) & 100.0 & 381 \\
AT-ResNet50-Linf ($\epsilon = 4/255$) & 100.0 & 243 \\
BagNet17 + PatchGuard & 55.2 & 4817 \\
\bottomrule
\end{tabular}

\end{table}

Despite the additional robustness mechanisms present in these models, as we see in Table ~\ref{tab:defender_results}, STROKE is still able to produce successful adversarial examples under the evaluated coverage-triggered black-box setting. As expected, patch-based defenses such as PatchGuard reduce the overall attack success rate and require substantially more queries compared to standard architectures. Nevertheless, the results suggest that sparse structural occlusions can still influence the predictions of these models, although the effectiveness varies depending on the underlying defense mechanism.

\subsection{Analysis of Learned Perturbations}

\subsubsection{Grad-CAM Visualization}

To qualitatively examine how STROKE perturbations influence model attention, we visualize Grad-CAM maps for several examples using ResNet50 under the standard evaluation setting (untargeted attack, nominal 2\% coverage threshold, and 10k-query limit). For each example, we show the clean image, the corresponding Grad-CAM map, the adversarial image generated by STROKE, and the Grad-CAM map of the adversarial prediction.

\begin{figure}[!t]
\centering
\includegraphics[width=\linewidth]{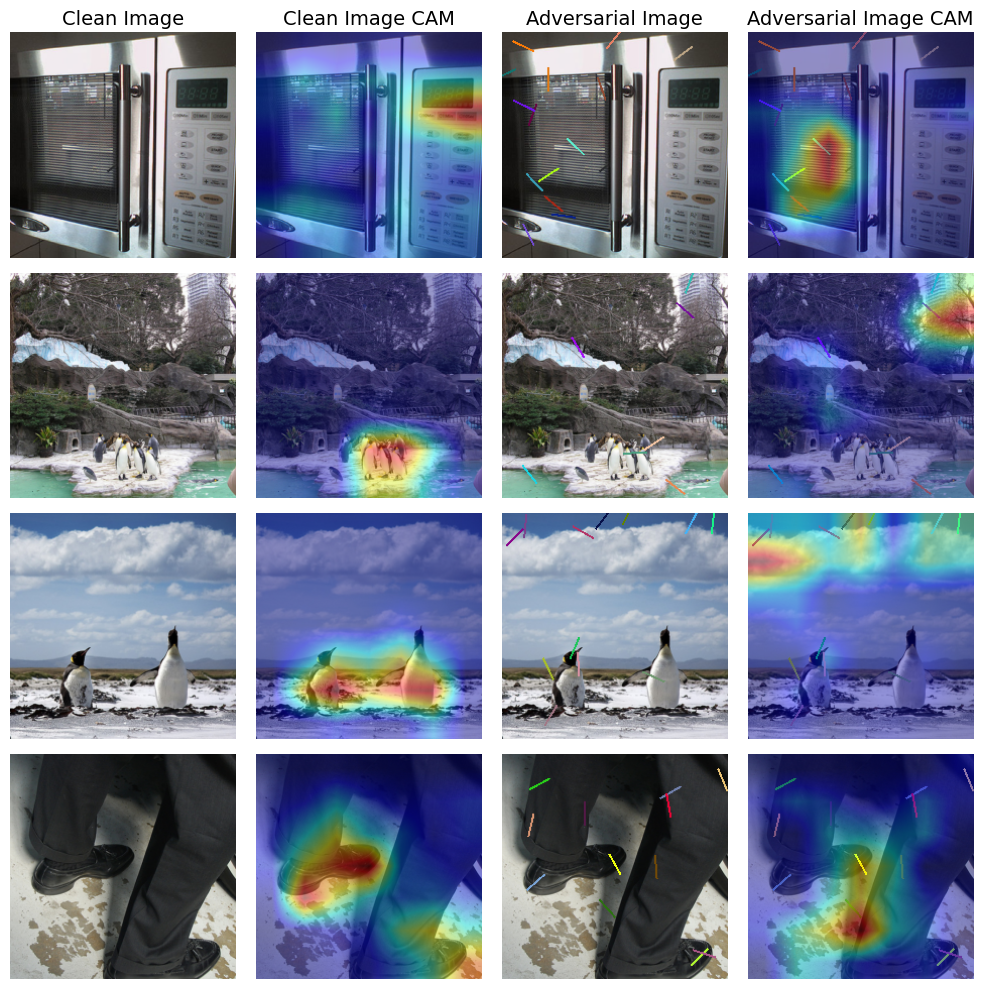}
\vspace{4pt}
\caption{
Grad-CAM visualizations for representative examples at the nominal 2\% pixel-coverage threshold and 10k-query limit on ResNet50. 
Columns show the clean image, the corresponding Grad-CAM map, the adversarial image produced by STROKE, and the Grad-CAM map of the adversarial prediction. 
The attention regions often shift away from the original object regions after the attack.
}
\label{fig:stroke_gradcam}
\end{figure}

As illustrated in Fig.~\ref{fig:stroke_gradcam}, the attention regions for the clean images generally align with the main object of interest. After applying the sparse occlusion perturbations, the Grad-CAM maps often shift toward different regions of the image. This behavior suggests that the structural occlusions introduced by STROKE can influence the model's attention patterns, even though the perturbations occupy only a small fraction of the image.

\subsubsection{t-SNE Feature Visualization}

To provide a qualitative view of how STROKE perturbations affect learned representations, we visualize feature embeddings of clean and adversarial images using t-SNE. Features are extracted from the penultimate layer of ResNet50 under the standard evaluation setting (untargeted attack, nominal 2\% coverage threshold, and 10k-query limit).

\begin{figure}[!t]
\centering
\includegraphics[width=0.9\linewidth,height=0.7\textheight,keepaspectratio]{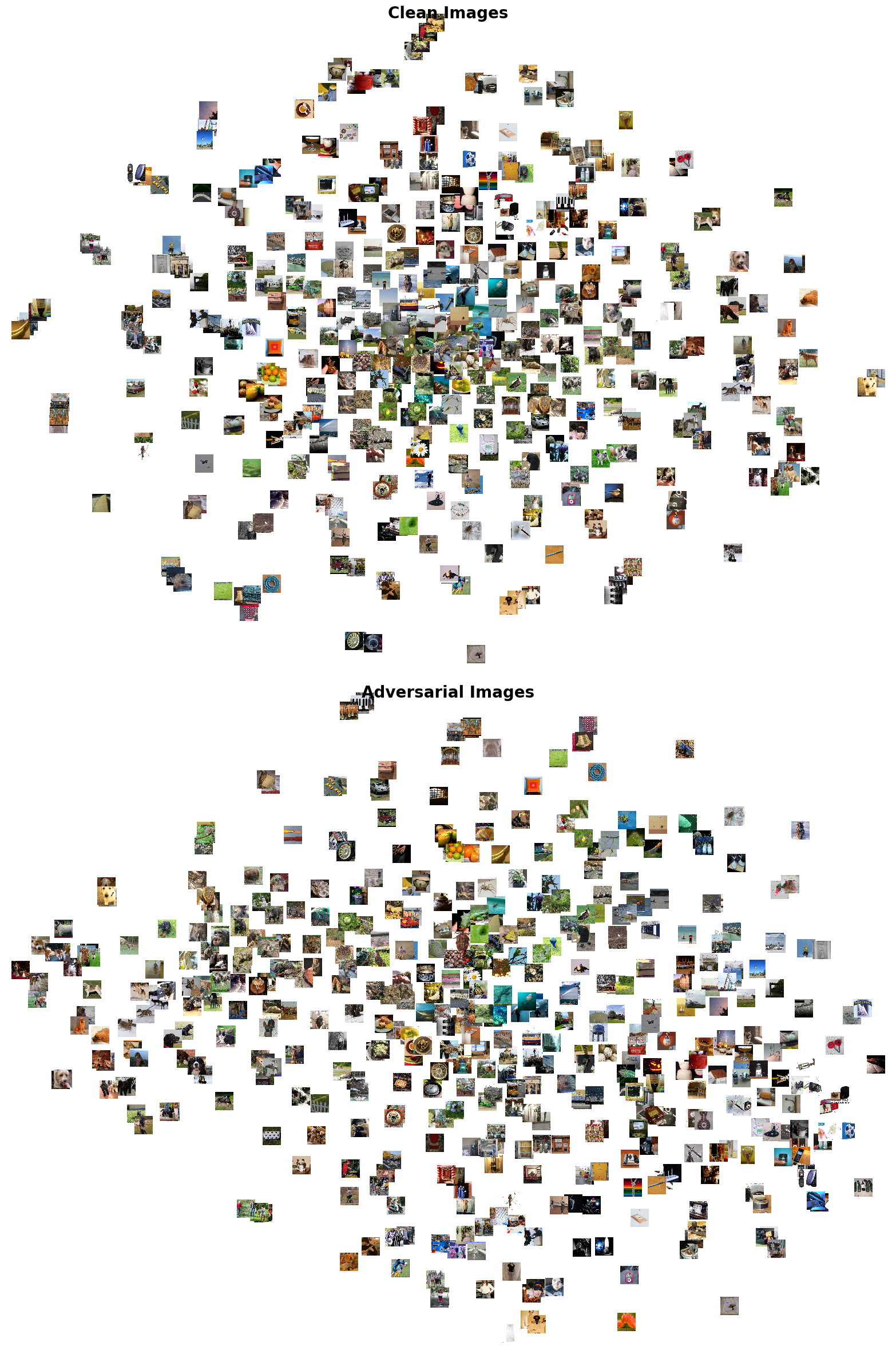}
\vspace{4pt}
\caption{
t-SNE visualization of ResNet50 feature embeddings for clean images (top) and adversarial images generated by STROKE (bottom). 
Each point is rendered as a thumbnail of the corresponding image. 
Adversarial perturbations lead to a noticeable redistribution of samples in the embedding space compared to the clean inputs.
}
\label{fig:tsne_stroke}
\end{figure}

Figure~\ref{fig:tsne_stroke} shows the resulting two-dimensional embeddings. Each point is rendered as a thumbnail of the corresponding image. While the clean images exhibit relatively coherent groupings in feature space, the embeddings of adversarial examples appear more dispersed, with several samples moving away from dense regions of the clean distribution.

Although t-SNE provides only a qualitative view of the feature space, this visualization suggests that the sparse structural perturbations introduced by STROKE can alter intermediate feature representations of the model.

\subsection{Transferability Experiments}

\subsubsection{Cross-Model Transferability}

We evaluate the transferability of adversarial examples generated by STROKE across different architectures. 
For each source model, adversarial examples are generated under the standard evaluation setting (untargeted attack, nominal 2\% coverage threshold, 10k-query limit), and then evaluated on other target models. 
The transfer attack success rate (ASR) is computed as the fraction of images for which the prediction of the target model changes between the clean and adversarial inputs.

\begin{figure}[!t]
\centering
\includegraphics[width=\linewidth]{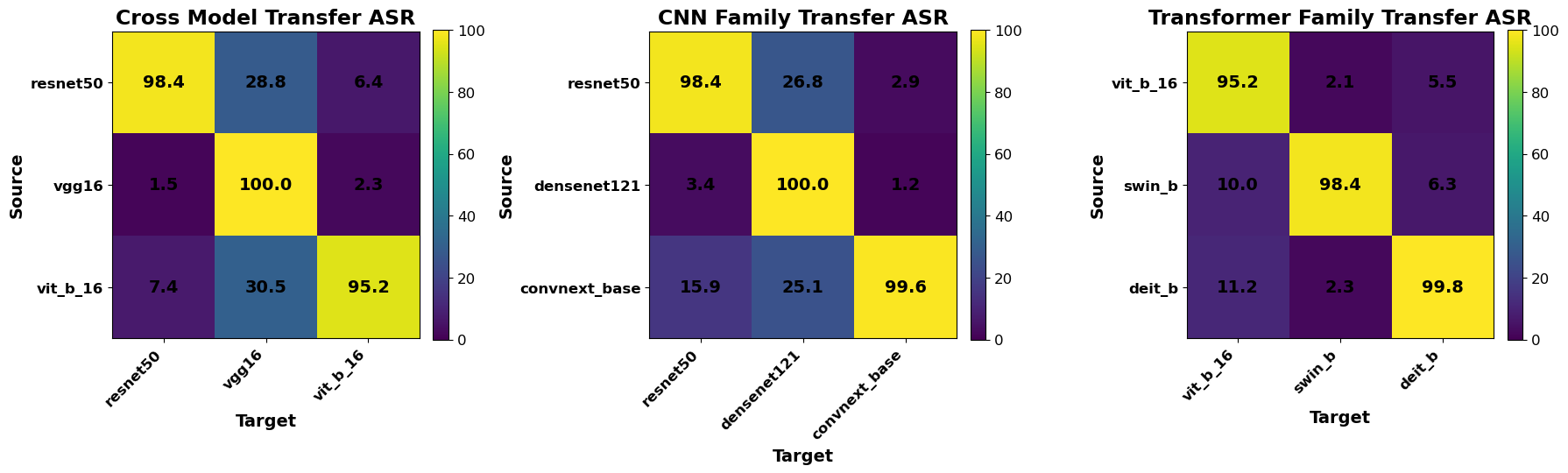}
\vspace{4pt}
\caption{
Transferability of adversarial examples generated by STROKE across different architectures. 
Rows correspond to the source model used to generate adversarial examples, and columns correspond to the target model used for evaluation. 
Each cell reports the attack success rate (ASR, \%). 
Results are shown for (left) cross-model transfer among commonly used architectures, (middle) transfer within CNN-based models, and (right) transfer within transformer-based models at a nominal 2\% coverage threshold and 10k-query limit.
}
\label{fig:stroke_transfer}
\end{figure}

Figure~\ref{fig:stroke_transfer} summarizes transfer performance across three settings: cross-model transfer between commonly used architectures, transfer within CNN families, and transfer within transformer-based models. Off-diagonal entries indicate varying degrees of transferability across architectures, with moderate transfer observed between related CNN models and comparatively lower transfer across CNN and transformer architectures.

\subsubsection{Physical Transfer}

To assess physical transferability, we conducted preliminary physical-world experiments using a custom web application hosted on a Pixel 7 smartphone. The application captured images with the device camera and performed real-time inference using a ResNet-50 model, returning the top-$N$ predictions. For consistency, all adversarial examples were photographed at a fixed distance of 20\,cm with no tilt.

Under this setup, STROKE achieved a digital untargeted attack success rate (ASR) of 99\% and a physical untargeted ASR of 46\%. For reference, IMPACT achieved a digital untargeted ASR of 85\% and a physical untargeted ASR of 52\% under a similar protocol. These results suggest that STROKE retains some degree of transfer after physical capture, although the success rate is lower than in the digital setting. 

\subsection{Additional Discussion}

The threat model considered in this work is motivated by naturalistic occlusions that commonly occur in real-world environments, such as small objects, markings, or localized artifacts appearing in the scene. 
Our results suggest that sparse structural perturbations can influence model predictions even when the affected area is small. 
Understanding such vulnerabilities may help inform the development of more robust visual recognition systems and encourage further investigation into defenses against localized perturbations.

\clearpage

\end{document}